\documentclass[letterpaper, 10 pt, conference]{ieeeconf}  

\IEEEoverridecommandlockouts                              

\usepackage{graphics} 
\usepackage{graphicx}
\usepackage{algorithmic}
\usepackage{amsmath} 
\usepackage{amssymb}  
\usepackage{cite}
\usepackage[hyphens]{url} 
\usepackage{amsfonts}   
\usepackage{booktabs}  
\usepackage{CJKutf8}
\usepackage{xcolor}
\usepackage{float}
\usepackage{stfloats} 

\usepackage{censor}
\usepackage{multirow}
\usepackage[ruled,vlined]{algorithm2e}

\usepackage{booktabs}

\usepackage{multirow}

\usepackage{makecell}

\title{\LARGE \bf TwinBridge: Digital Twin-Driven Policy and Reward Transfer for Real-World Reinforcement Learning in Laboratory Robotics}

\author{
  Shu'ang Sun\textsuperscript{$\dagger$},
  Yibo Qiu\textsuperscript{$\dagger$},
  Haoliang Ye,
  Zan Huang,\\
  Ao Tian,
  Kan Ding,
  Ronald Xu,
  Mingzhai Sun\textsuperscript{$\ast$}%
  \thanks{$^\dagger$Equal contribution. $^\ast$Corresponding author.}%
  \thanks{All authors are with the Suzhou Institute for Advanced Research, University of Science and Technology of China, Suzhou, Jiangsu, China, and also with the School of Biomedical Engineering, Division of Life Sciences and Medicine, University of Science and Technology of China, Hefei, Anhui, China.}%
  \thanks{Emails: \{sunshuang,alexandreqiu,huangzan,dingkan\}@mail.ustc.edu.cn,
    15005704869@163.com (H. Ye), 3411896198@qq.com (A. Tian),
    \{xux,mingzhai\}@ustc.edu.cn.}%
  \thanks{This work was supported by the Gusu Leading Talent Entrepreneurship and Innovation Program (Grant No. ZL2024349).}%
  \thanks{This work has been submitted to the IEEE for possible publication. Copyright may be transferred without notice, after which this version may no longer be accessible.}%
}

\DeclareGraphicsExtensions{.png}
\newcommand{\paperfigure}[3]{%
  \IfFileExists{#1.png}{\includegraphics[width=\linewidth]{#1.png}}{%
    \fbox{\begin{minipage}[c][#2][c]{\dimexpr\linewidth-2\fboxsep-2\fboxrule\relax}
      \centering\small\textbf{#3}\par\medskip
      Figure to be added.\par\smallskip
      \footnotesize\texttt{\detokenize{#1}.png}
    \end{minipage}}%
  }%
}

\begin{document}
\bstctlcite{IEEEBSTControl}
\begin{CJK*}{UTF8}{gbsn}
\maketitle
\thispagestyle{empty}
\pagestyle{empty}

\begin{abstract}
Online reinforcement learning offers an effective approach to precise laboratory manipulation, yet costly real-world demonstrations and limited access to dense process rewards hinder its practical application. We present TwinBridge, a digital twin-driven framework that supplies policy initialization and dense visual rewards for real-world online learning. We construct digital twins by combining 3D Gaussian splatting with physics simulation. Here, simulator states provide task events and geometric progress under physical prerequisites and ordering constraints, yielding rewards for policy pretraining with simulated demonstrations and reinforcement learning. We distill this supervision into a visual reward model. During real-world interaction, its predictions are combined with execution history to compute dense rewards for human-in-the-loop policy refinement. Across five real-world laboratory tasks, TwinBridge achieves a final mean autonomous success rate of 98\% with at most 25 minutes of real-world interaction per task. Experiments show that dense rewards improve interaction efficiency over sparse rewards, and simulation-based policy initialization further accelerates learning.
\end{abstract}

\section{INTRODUCTION}
\begin{figure*}[t]
  \centering
  \includegraphics[width=\linewidth]{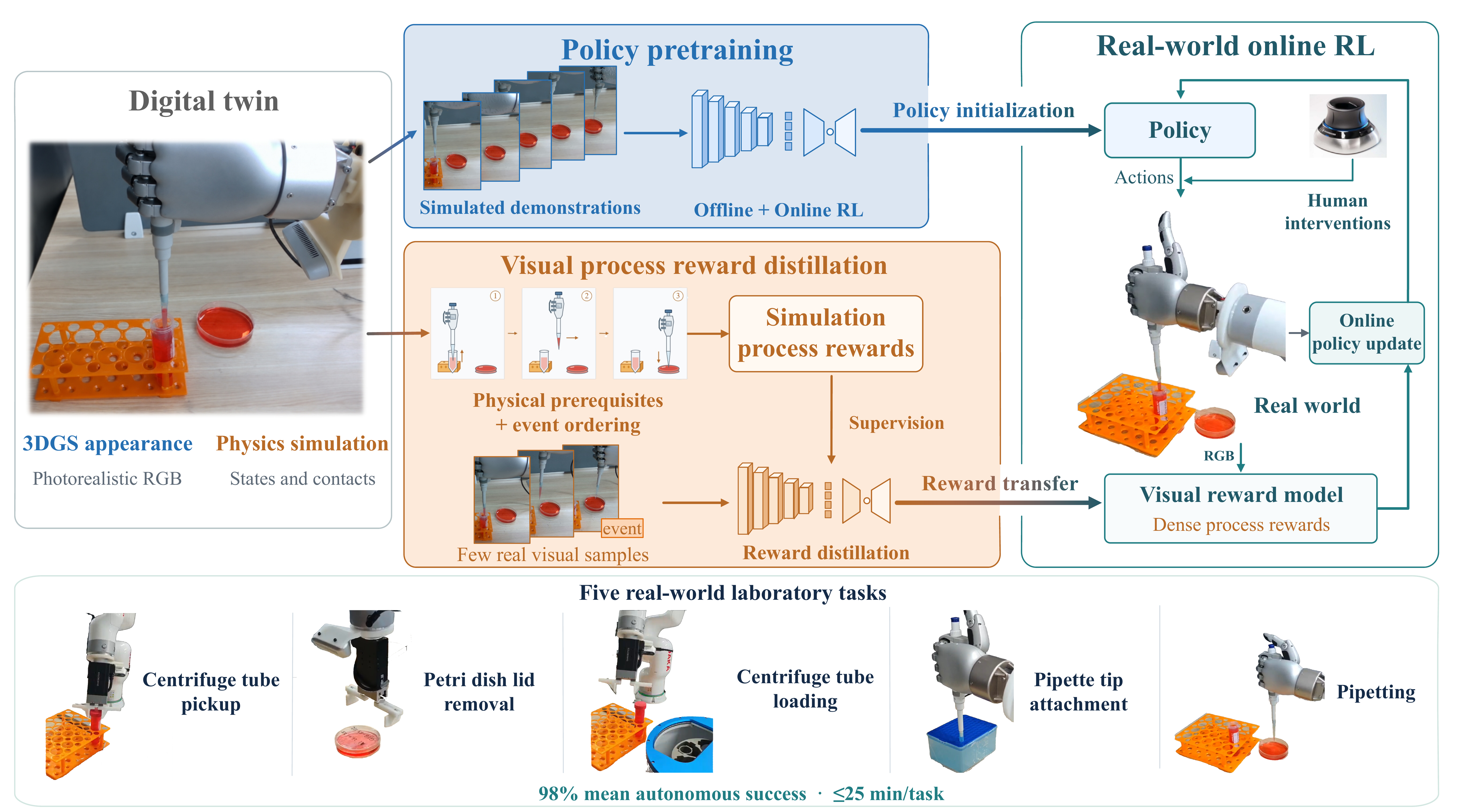}
  \caption{Overview of TwinBridge. A digital twin supplies two resources for real-world online RL: a policy pretrained with simulated demonstrations and RL, and visual process rewards learned from physical supervision and a small set of real annotations. During real interaction, the frozen reward model guides policy refinement with human interventions. The bottom row shows the five laboratory tasks.}
  \label{fig:twinbridge_motivation}
\end{figure*}

Laboratory manipulation requires precise motion while satisfying physical dependencies between successive actions. A centrifuge tube must be securely grasped before transport, and a lid must be separated from its container before relocation. Reinforcement learning (RL) can improve these skills through interaction, but efficient learning on a physical robot requires both useful initial behavior and feedback that identifies valid intermediate progress.

Human-in-the-loop methods such as HIL-SERL and HIL-ConRFT have demonstrated precise manipulation by combining demonstrations, human corrections, and reinforcement learning~\cite{luo2025precise,chen2025conrft}. For multi-stage laboratory tasks, collecting demonstrations that cover successive operations requires skilled human effort. Sparse outcome feedback further leaves the learner to discover which intermediate actions advance the procedure.

Digital twins allow useful behavior to be learned before real-world interaction. SplatSim and GSWorld combine Gaussian splatting with physics simulation to support visual policy transfer~\cite{qureshi2025splatsim,jiang2026gsworld}. SimLauncher and TwinRL further use simulation experience to guide subsequent real-world RL~\cite{wu2025simlauncher,xu2026twinrl}. Simulation thus supports both policy preparation and continued learning after transfer. We focus on a complementary opportunity: transferring the task-progress information available in simulation alongside the policy, so that the twin supplies both initial behavior and feedback for real-world improvement.

Visual reward modeling provides a way to deliver such feedback from camera observations. SARM models task stages and within-stage progress~\cite{chen2026sarm}, Robo-Dopamine uses visual process rewards for RL~\cite{tan2026robodopamine}, and keyframe-guided structured rewards exploit task order in laboratory manipulation~\cite{qiu2026keyframe}. In precision manipulation, progress also depends on physical prerequisites: moving the gripper toward a holder advances tube loading only after a grasp is secured. A digital twin exposes contact, alignment, and insertion states, allowing valid events and geometric progress to be labeled automatically. The central challenge is to transfer this supervision to visual predictions while retaining event order and execution history in the deployed reward computation.

We propose \textbf{TwinBridge}, a digital twin-driven framework that supplies pretrained policies and dense visual process rewards for real-world online learning (Fig.~\ref{fig:twinbridge_motivation}). We combine 3D Gaussian splatting with physics simulation to construct digital twins, where simulator states provide task events and geometric progress under physical prerequisites and ordering constraints. These signals define rewards for visual policy pretraining with simulated demonstrations and RL, and provide supervision aligned with simulated images. We combine this supervision with a small set of real visual annotations to train a visual reward model. During real-world interaction, its event and progress predictions are combined with execution history to compute dense rewards, guiding human-in-the-loop online refinement of the transferred policy. An execution-history ledger credits each event once and rewards new valid progress records.

We evaluate TwinBridge on five laboratory tasks: centrifuge tube pickup, Petri dish lid removal, centrifuge tube loading, pipette tip attachment, and pipetting. TwinBridge achieves a final mean autonomous success rate of 98\% with at most 25 minutes of real-world interaction per task. Dense visual rewards improve interaction efficiency over sparse outcome feedback, while simulation-based initialization further accelerates real-world learning.

The main contributions of this work are:
\begin{itemize}
    \item A digital twin-driven framework that jointly transfers visual manipulation policies and process reward supervision to support efficient human-in-the-loop real-world RL.
    \item A physically grounded reward construction and visual distillation method that combines ordered event credit and valid progress records with execution history to provide dense feedback from real-world images.
    \item Evaluation on five real-world laboratory tasks, including comparisons of policy initialization and reward-transfer configurations and episode-level validation of visual event and progress prediction.
\end{itemize}

\section{RELATED WORK}
\label{sec:related_work}
\begin{figure*}[t]
  \centering
  \includegraphics[width=\linewidth]{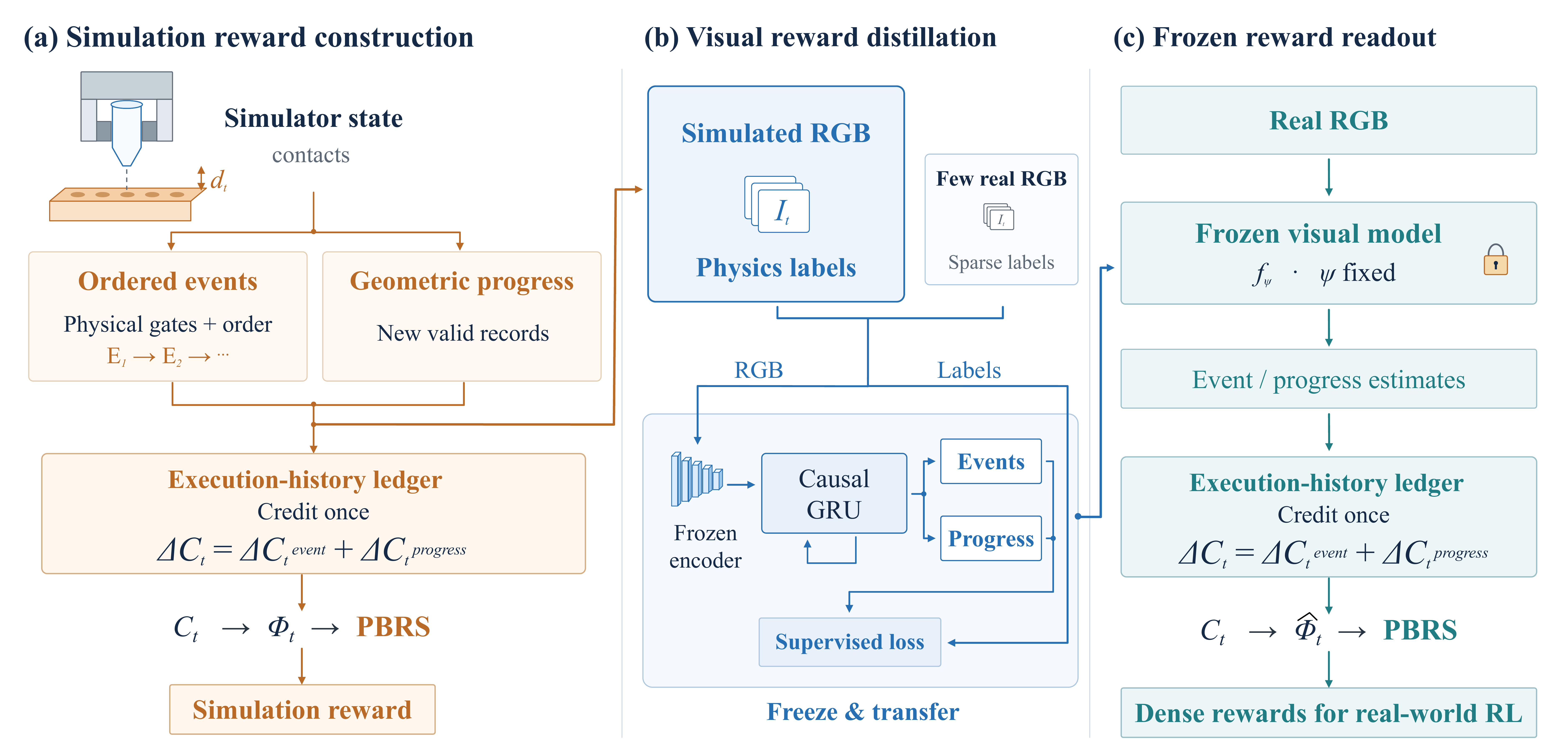}
  \caption{Physics-grounded reward construction and visual transfer.
  (a) Ordered events and new valid progress records receive one-time credit in an execution-history ledger, defining the potential for potential-based reward shaping (PBRS).
  (b) Labeled simulated RGB sequences and a small set of real RGB sequences with event/stage annotations train a causal visual predictor.
  The trained predictor is frozen and transferred.
  (c) Real RGB predictions update the visual ledger under the same ordering and progress-record rules, and the same PBRS readout supplies online process rewards.}
  \label{fig:twinbridge_overview}
\end{figure*}

\subsection{Laboratory Automation and Real-World Reinforcement Learning}

Laboratory automation connects experimental planning with physical execution. Mobile robotic chemists coordinate instruments and experiments~\cite{burger2020chemist,dai2024autonomous}, while Coscientist and BioMARS integrate language-based planning with laboratory tools and robotic workflows~\cite{boiko2023autonomous,qiu2025biomars}. Reliable grasping, lid handling, and insertion policies provide the manipulation capabilities underlying these workflows. Real-world RL improves such skills through interaction~\cite{ibarz2021robot,kalashnikov2018qtopt}.

Prior data and human guidance improve learning efficiency. RLPD integrates offline data into online RL~\cite{ball2023rlpd}, and IBRL uses an imitation policy for exploration and value bootstrapping~\cite{hu2024ibrl}. HG-DAgger aggregates expert corrections for imitation learning~\cite{kelly2019hgdagger}. HIL-SERL combines demonstrations, interventions, and visual success classifiers for precise manipulation~\cite{luo2025precise}; HIL-ConRFT connects offline behavioral supervision and value learning with online reinforcement fine-tuning of pretrained policies~\cite{chen2025conrft}. GRAFT uses region-level visual supervision to guide efficient online adaptation for fine-grained biomedical manipulation~\cite{qiu2026graft}. TwinBridge builds on human-in-the-loop RL, using simulation to supply both an initial manipulation policy and process rewards for subsequent real-world interaction.

\subsection{Digital Twins for Policy Learning}

Sim-to-real methods address discrepancies in appearance and dynamics through visual domain randomization~\cite{tobin2017randomization} and dynamics randomization~\cite{peng2018dynamics}. Reconstructed digital twins provide training environments corresponding to physical scenes. RialTo improves real-world imitation policies through RL in reconstructed twins~\cite{torne2024rialto}. SplatSim and RE$^3$SIM combine scene reconstruction and photorealistic rendering to generate transferable manipulation data~\cite{qureshi2025splatsim,han2026re3sim}. GSWorld couples Gaussian scene assets with physics for policy training, evaluation, and corrective data collection~\cite{jiang2026gsworld}. RL-GSBridge couples Gaussian rendering with physics for visual RL transfer~\cite{wu2025rlgsbridge}, while TwinAligner aligns appearance and dynamics using reconstruction and physical parameter identification~\cite{fan2025twinaligner}.

Simulation-trained policies can also be adapted through real-world interaction. TRANSIC learns residual policies from human corrections to bridge sim-to-real gaps~\cite{jiang2025transic}. SimLauncher uses simulation-pretrained action proposals and simulated demonstrations to guide real-world exploration and value learning~\cite{wu2025simlauncher}. TwinRL transfers twin-generated experience and guides real-world exploration~\cite{xu2026twinrl}. SGFT uses a simulation-trained value function for reward shaping and shorter-horizon adaptation~\cite{yin2025sgft}. TwinBridge transfers both a manipulation policy and simulator-derived supervision for visual process rewards.

\subsection{Visual Process Rewards}

Visual reward learning extracts intermediate feedback from observations. Related work estimates activity progress using representations learned through self-supervised video alignment~\cite{donahue2024progress}. XIRL learns progress-sensitive representations through temporal alignment across embodiments~\cite{zakka2022xirl}; VIP and LIV learn visual and language-conditioned rewards from passive videos~\cite{ma2023vip,ma2023liv}. RL-VLM-F learns rewards from vision-language model preferences~\cite{wang2024rlvlmf}, and RoboReward trains reward models on large robot datasets with negative and partial-progress examples~\cite{lee2026roboreward}. ReWiND transfers language-conditioned rewards to policy learning for new tasks~\cite{zhang2025rewind}. TimeRewarder estimates frame-pair temporal distances to provide step-wise progress feedback~\cite{liu2026timerewarder}.

Task structure also informs reward construction: reward machines represent event-dependent rewards with finite-state models~\cite{icarte2018rewardmachines}. SARM derives stage-aware progress targets from subtask annotations for demonstration filtering and reweighting~\cite{chen2026sarm}. ProcVLM grounds progress in procedural structure and intra-stage visual change~\cite{feng2026procvlm}. Robo-Dopamine combines local and global progress estimates with potential-based shaping for RL~\cite{tan2026robodopamine}. Keyframe-guided structured rewards compare observations with demonstration-derived targets and use ordered stages with one-time advancement credit for laboratory manipulation~\cite{qiu2026keyframe}.

TwinBridge grounds process supervision in simulator states, with physical prerequisites and event order defining valid advancement. Simulation-derived labels and a small set of real visual annotations train an event and progress predictor. During real-world learning, its predictions update an execution-history ledger that credits new events and eligible progress records. A potential-based readout~\cite{ng1999shaping} converts these updates into dense rewards, carrying the simulated task structure into human-in-the-loop policy refinement.

\section{METHOD}
\label{sec:method}

TwinBridge prepares a manipulation policy and visual process rewards in a digital twin, then transfers both to human-in-the-loop real-world RL. Figure~\ref{fig:twinbridge_overview} traces the reward pathway from simulation reward construction (a), through visual distillation (b), to frozen real-world reward readout (c). Algorithm~\ref{alg:twinbridge} summarizes training and deployment.

\subsection{Digital Twin and Transfer Pipeline}
\label{sec:task_formulation}

We couple physics simulation with 3D Gaussian splatting~\cite{kerbl2023gaussian}. Rigid-body poses update the corresponding Gaussians, synchronizing rendered RGB observations with robot--object interactions. The policy receives external and wrist images, proprioception, and a task instruction; it outputs relative Cartesian motion and a task-specific tool command. Simulated demonstrations and subsequent RL provide policy initialization. In parallel, simulator contact, alignment, and insertion states supply supervision for the visual reward model. Figure~\ref{fig:real_twin} shows the tasks and their paired real and rendered observations.

\subsection{Physics-Grounded Process Rewards}
\label{sec:privileged_reward}

Each task defines ordered events $\mathcal E=(e_1,\ldots,e_K)$ and their physical prerequisites. For tube loading, transport requires a secured grasp and rack clearance; insertion additionally requires alignment. An execution-history ledger stores completed events, accumulated credit $C_t$, and the highest valid geometric progress $G_t$. New events receive credit once, after their prerequisites and preceding events are satisfied.

For a nonterminal transition, let $\mathcal N_t$ denote newly confirmed events and $g_{t+1}\in[0,1]$ the successor progress. Credit accumulates as
\begin{equation}
 C_{t+1}=C_t+B_{\mathrm{evt}}\sum_{k\in\mathcal N_t}w_k
                 +\Delta C_t^{\mathrm{prog}},
 \label{eq:credit_update}
\end{equation}
where $B_{\mathrm{evt}}$ is the event budget and $w_k$ are normalized task-specific weights. If the successor satisfies the physical eligibility conditions and $g_{t+1}>G_t+\delta$, it establishes a new progress record, updates $G_t$, and adds $\Delta C_t^{\mathrm{prog}}=\epsilon_p\gamma^t$; otherwise the increment is zero. Here $\delta$ is the record tolerance and $\epsilon_p\geq0$ is the task-configured progress coefficient. Thus progress credit depends on new valid records, while event credit marks discrete milestones.
For tube loading, clearing the rack marks a discrete milestone, while subsequent eligible improvements in alignment or insertion depth can establish progress records.
The two credit terms therefore provide feedback at complementary resolutions: task events and valid geometric advancement between events.

With $C_0=0$, $G_0=g_0$, and discount $\gamma$, the potential is
\begin{equation}
 \Phi_t=\begin{cases}
 C_t/\gamma^t,&\text{within the episode},\\
 0,&\text{at an absorbing episode end}.
 \end{cases}
 \label{eq:ledger_potential}
\end{equation}
In this construction, $g_t$ represents current progress evidence, $C_t$ stores credited advancement, and $\Phi_t$ encodes that accumulated history for reward shaping.
We apply potential-based reward shaping (PBRS) to this history- and time-dependent reward state~\cite{ng1999shaping,devlin2012dynamic}:
\begin{equation}
 r_t=r_t^{\mathrm{base}}
      +\beta(\gamma\Phi_{t+1}-\Phi_t)+r_t^{\mathrm{aux}}.
 \label{eq:total_reward}
\end{equation}
Here $\beta$ scales shaping. The base term encodes task outcomes and step cost; the auxiliary term penalizes configured invalid actions. On nonterminal transitions, shaping equals $\beta(C_{t+1}-C_t)/\gamma^t$. Repeated or regressed progress earns no new credit, although the potential retains execution history.
For example, retracting and reinserting a tube to an already credited depth does not establish a new progress record. Success, failure, the task's finite horizon, and finalized aborts set the successor potential to zero. Ordinary sequence boundaries preserve it, distinguishing task termination from training truncation~\cite{pardo2018timelimits}.

\subsection{Visual Reward Distillation and Frozen Readout}
\label{sec:visual_reward}

The physical reward construction supplies frame-aligned event and geometric targets for simulated RGB sequences.
The simulated image and its targets are generated from the same underlying state, preserving their correspondence throughout object motion and contact transitions. Real visual annotations specify event states or transition intervals, with uncertain boundaries masked.
Real annotations associate these task milestones with real visual observations, providing supervision for transfer across the appearance gap.

A frozen Octo encoder~\cite{team2024octo} produces visual features $z_t$. The distilled student model, hereafter the visual predictor, uses gated recurrent units~\cite{cho2014gru} to causally estimate event probabilities $\widehat{\mathbf e}_t$ and progress $\widehat g_t$:
\begin{equation}
 (h_t,\widehat{\mathbf e}_t,\widehat g_t)
 =f_\psi(z_t,\Delta z_t,h_{t-1}),
 \label{eq:visual_prediction}
\end{equation}
where $h_t$ is recurrent state and $\Delta z_t$ captures visual change.

Offline training combines the available supervision in each domain:
\begin{equation}
 \mathcal L_{\mathrm{vis}}
 =\mathbb E_{\mathcal D_{\mathrm{real}}}[\ell_{\mathrm{obs}}]
 +\alpha_{\mathrm{sim}}\mathbb E_{\mathcal D_{\mathrm{sim}}}[\ell_{\mathrm{obs}}].
 \label{eq:visual_loss}
\end{equation}
The loss combines masked, class-balanced event classification, stage classification, simulator progress regression, stage-derived progress intervals, and order/temporal consistency terms.
Real annotations anchor event and stage predictions to real visual observations and impose coarse interval constraints on progress. They do not provide framewise geometric progress targets within a stage. Direct geometric progress regression uses simulator-derived targets wherever available, with supervision masked according to label availability.

During real-world learning, the visual predictor supplies evidence with fixed parameters $\psi$ while updating its recurrent state.
The ordered readout confirms predicted events under the task's ordering rule and updates the visual ledger with new event credit and eligible progress records. Equation~\eqref{eq:ledger_potential} gives $\widehat\Phi_t$, and Eq.~\eqref{eq:total_reward} produces the reward.

\subsection{Policy Transfer and Online Refinement}
\label{sec:policy_transfer}
\label{sec:online_optimization}

We use HIL-ConRFT's actor--critic optimization~\cite{chen2025conrft}, with frozen Octo representations and a consistency-model action head~\cite{song2023consistency}. The simulation-trained actor initializes real-world learning, where an operator can replace policy actions. Replay stores executed actions, visual rewards, successor observations, terminal masks, and intervention information. Critic updates minimize Bellman error, while actor updates combine value optimization with corrective supervision. The visual ledger augments critic inputs; the actor retains its original observations. The frozen reward model therefore supplies process feedback while the policy and critics adapt through real interaction.

\section{EXPERIMENTS}
\label{sec:experiments}
\raggedbottom
\setcounter{dbltopnumber}{1}
\setcounter{topnumber}{1}
\begin{figure*}[t]
  \centering
  \includegraphics[width=\linewidth]{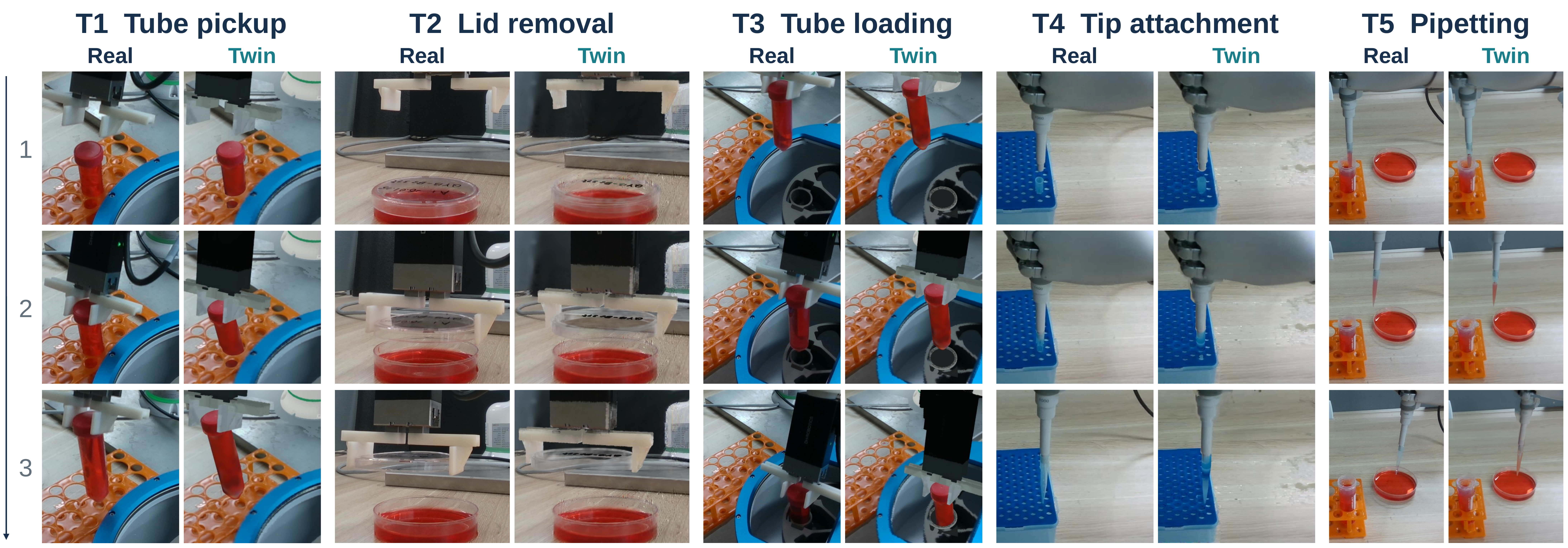}
  \caption{Five laboratory tasks across four scenes. T1--T5 show tube pickup, lid removal, tube loading, tip attachment, and pipetting; T1 and T3 share a scene. Three representative stages run from top to bottom. Each stage pairs a real observation (left) with its digital-twin render (right).}
  \label{fig:real_twin}
\end{figure*}

We evaluated learning in digital twins, real-world policy refinement, and the event and progress feedback supplied by the visual reward model.

\begin{figure}[!b]
  \centering
  \includegraphics[width=\columnwidth]{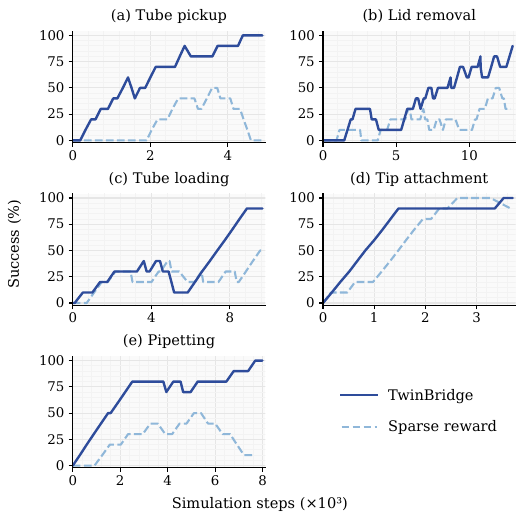}
  \caption{Simulation online learning across five laboratory tasks. Curves show ten-episode rolling success rates for TwinBridge and sparse-reward baseline, plotted against simulation steps. Both methods are displayed over the same interaction budget within each task.}
  \label{fig:sim_pretraining}
\end{figure}

\begingroup
\makeatletter\def\@secpenalty{0}\makeatother
\subsection{Experimental Platform and Tasks}
\label{sec:experimental_setup}

The platform used a JAKA Minicobo with a parallel gripper for tube and lid manipulation, and a LinkerHand L7 holding an Eppendorf single-channel pipette for the pipette tasks. Simulation pretraining and real-world online learning used NVIDIA RTX 5880 Ada GPUs (48~GB); offline reward-model evaluation used a single L40S or RTX 5880 Ada GPU (48~GB) per run. Isaac Sim 4.5 provided physics simulation, with Gaussian splatting for RGB rendering.
\par\endgroup

External and wrist images had resolutions of $256\times256$ and $128\times128$, respectively. The policy also received two frames of 19-dimensional proprioception. Cartesian control ran at 10~Hz; episodes lasted at most 512 steps, with the initial TCP position randomized within $\pm 20$~mm along both horizontal axes.

Figure~\ref{fig:real_twin} shows five tasks across four scenes. \textit{Tube pickup} lifts a centrifuge tube clear of its rack; \textit{lid removal} separates and relocates a Petri dish lid; \textit{tube loading} transfers a tube from its rack into the centrifuge holder; \textit{tip attachment} attaches and lifts a pipette tip; and \textit{pipetting} comprises aspiration, loaded transfer, and dispensing. Tube pickup and tube loading share a scene.

\subsection{Baselines and Evaluation Protocol}
\label{sec:evaluation_protocol}

We evaluated the two transfer pathways within the ConRFT framework~\cite{chen2025conrft}. TwinBridge \textit{w/ sim policy} and \textit{w/o sim policy} compare learning with and without simulation policy initialization under frozen visual process rewards. TwinBridge \textit{w/o sim policy} and HIL-ConRFT compare visual-process and sparse-outcome reward configurations without simulation policy initialization. We also evaluated HG-DAgger~\cite{kelly2019hgdagger}, which learns from human corrections, and HIL-SERL~\cite{luo2025precise}, which uses sparse outcome feedback.

\begin{figure*}[t]
  \centering
  \includegraphics[width=\linewidth]{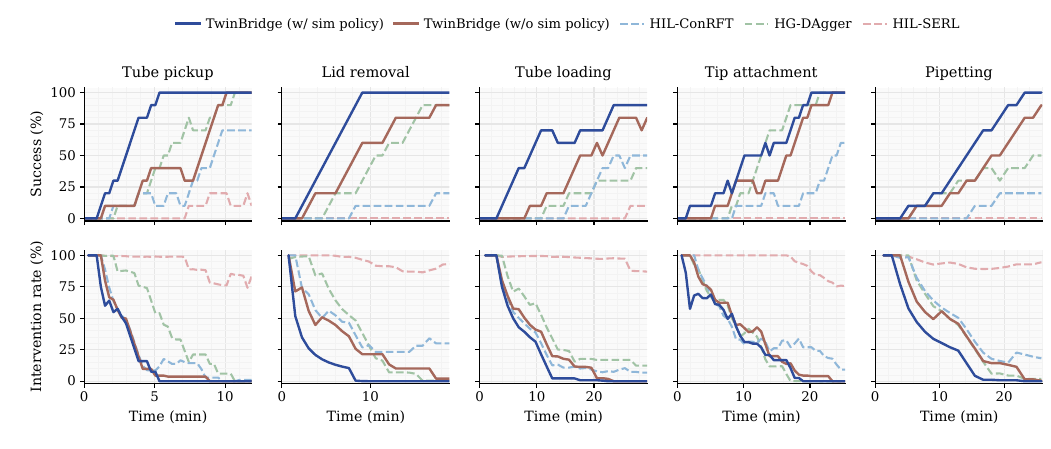}
  \caption{Real-world online learning. Columns follow the task order in Fig.~\ref{fig:real_twin}. Ten-episode rolling averages show task success (top) and episode-level intervention fractions (bottom). Both TwinBridge variants use frozen visual process rewards; w/ and w/o sim policy denote the presence and absence of simulation policy initialization.}
  \label{fig:real_finetuning}
\end{figure*}

Figure~\ref{fig:real_finetuning} reports recorded success and operator-controlled action fractions with ten-episode rolling windows. The time axis reports real-world online interaction time. Table~\ref{tab:real_results} reports final online success rates and their equally weighted mean across tasks. Autonomous success requires task completion without operator-controlled actions.

\begin{table}[t]
  \centering
  \small
  \caption{Final Online Success Rates (\%)}
  \label{tab:real_results}
  \setlength{\tabcolsep}{1pt}
  \renewcommand{\arraystretch}{1.08}
  \resizebox{\columnwidth}{!}{%
  \begin{tabular}{@{}cccccc@{}}
    \toprule
    \multirow{2}{*}{Task}
      & \multicolumn{2}{c}{TwinBridge}
      & \multirow{2}{*}{\makecell{HIL-\\ConRFT}}
      & \multirow{2}{*}{\makecell{HG-\\DAgger}}
      & \multirow{2}{*}{\makecell{HIL-\\SERL}} \\
    \cmidrule(lr){2-3}
      & w/ sim policy & w/o sim policy & & & \\
    \midrule
    Tube pickup         & \textbf{100} & \textbf{100} & 70 & \textbf{100} & 10 \\
    Lid removal      & \textbf{100} & 90           & 20 & 90           & 0  \\
    Tube loading          & \textbf{90}  & 80           & 50 & 40           & 10 \\
    Tip attachment    & \textbf{100} & \textbf{100} & 60 & \textbf{100} & 0  \\
    Pipetting        & \textbf{100} & 90           & 20 & 50           & 0  \\
    \midrule
    Mean             & \textbf{98}  & 92           & 44 & 76           & 4  \\
    \bottomrule
  \end{tabular}}
  \par\smallskip
  \begin{minipage}{\columnwidth}
    \footnotesize
    Success rates are computed over the final ten online episodes of each reported training interval. w/ sim policy and w/o sim policy denote the variants with and without simulation policy initialization; both use visual process rewards. Mean denotes the equally weighted average across tasks; bold marks row-wise best results, including ties.
  \end{minipage}
\end{table}

\newsavebox{\TwinBridgeRewardReadout}
\begin{lrbox}{\TwinBridgeRewardReadout}
\begin{minipage}{\columnwidth}
  \makeatletter\def\@captype{figure}\makeatother
  \centering
  \includegraphics[width=\columnwidth]{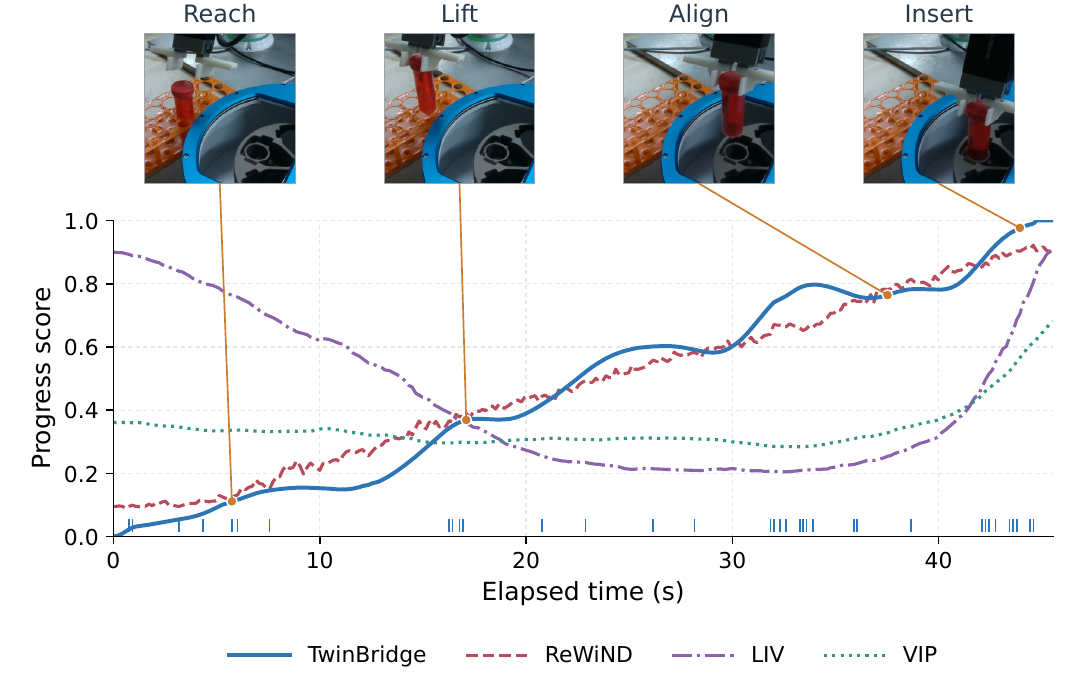}
  \caption{Visual progress scores during tube loading. LIV and VIP derive progress scores by comparing each frame with the final frames of training demonstrations. Images show key stages along the TwinBridge curve. Blue ticks mark progress-credit updates in the TwinBridge ledger.}
  \label{fig:visual_rewards}
\end{minipage}
\end{lrbox}

\begin{figure}[t]
  \centering
  \usebox{\TwinBridgeRewardReadout}
\end{figure}

\subsection{Online Learning in Digital Twins}
\label{sec:sim_pretraining}

Figure~\ref{fig:sim_pretraining} compares TwinBridge and sparse-reward learning in digital twins. Paired runs used matched offline update budgets and separately pretrained policies, retaining their respective reward settings. Within each shared displayed interaction budget, TwinBridge reached 100\%, 90\%, and 90\% rolling success on tube pickup, lid removal, and tube loading, whereas the sparse baseline reached at most 50\%. On pipetting, TwinBridge reached 100\% while the sparse baseline peaked at 50\%; both methods reached 100\% on tip attachment.

\subsection{Real-World Online Refinement}
\label{sec:real_finetuning}

TwinBridge achieved 100\% final success on four tasks and 90\% on tube loading, yielding a mean of 98\% (Table~\ref{tab:real_results}). These final episodes involved no human intervention.

Without simulation policy initialization, TwinBridge achieved 92\% mean final online success, compared with 44\% for HIL-ConRFT. The gains span all five tasks, ranging from 30 to 70 percentage points.

HG-DAgger and HIL-SERL achieved mean success rates of 76\% and 4\%, respectively. Relative to HG-DAgger, TwinBridge increased final success by 10, 50, and 50 percentage points on lid removal, tube loading, and pipetting, respectively.

Simulation initialization also improved learning speed.
In Fig.~\ref{fig:real_finetuning}, the \textit{w/ sim policy} variant first reached 80\% rolling success earlier than the \textit{w/o sim policy} variant on every task. For tube pickup and lid removal, the corresponding times were approximately 3.9 versus 9.2 minutes and 7.6 versus 12.9 minutes.
Both variants reached 100\% on tube pickup and tip attachment, while simulation initialization accelerated learning.

The intervention curves show how human assistance declined as the policies improved. Together, the success and intervention measurements connect transferred initialization to faster acquisition of autonomous behavior.

\begingroup
\setlength{\intextsep}{4pt}
\begin{table}[t]
  \setlength{\abovecaptionskip}{4pt}
  \centering
  \caption{Visual Reward Model Evaluation on Real Trajectories}
  \label{tab:visual_reward_accuracy}
  \small
  \setlength{\tabcolsep}{4pt}
  \renewcommand{\arraystretch}{1.04}
  \begin{tabular*}{\columnwidth}{@{\extracolsep{\fill}}ccc@{}}
    \toprule
    Task & Event F1 (\%) $\uparrow$ & Temporal $\rho$ $\uparrow$ \\
    \midrule
    Tube pickup & 99.50 & 0.925 \\
    Lid removal & 99.61 & 0.881 \\
    Tube loading & 98.20 & 0.893 \\
    Tip attachment & 99.51 & 0.882 \\
    Pipetting & 99.69 & 0.862 \\
    \midrule
    Mean & \textbf{99.30} & \textbf{0.889} \\
    \bottomrule
  \end{tabular*}
  \par\smallskip
  \begin{minipage}{\columnwidth}\footnotesize
    Predictions on held-out successful real trajectories. Event F1 uses threshold 0.7; temporal $\rho$ is the mean per-trajectory Spearman correlation of visual progress with elapsed time.
  \end{minipage}
\end{table}

\endgroup

\subsection{Visual Reward Quality and Readout}
\label{sec:reward_analysis}

For Table~\ref{tab:visual_reward_accuracy}, we evaluated visual predictors on 50 successful real trajectories using five episode-level folds: eight training and two test trajectories per task and fold. Each TwinBridge visual predictor received 40 simulation episodes and its training-fold real annotations for 2,000 updates. Event F1 pools known frame--event labels at threshold 0.7. Following ReWiND~\cite{zhang2025rewind}, temporal $\rho$ measures Spearman correlation between raw visual progress and elapsed time per complete trajectory, then averages trajectories within each task. Predictions precede ledger readout. The task means were 99.30\% F1 and 0.889 temporal $\rho$, supporting event recognition and temporally aligned progress estimation on held-out real observations.
Event F1 assesses milestone detection by balancing precision and recall, whereas temporal $\rho$ assesses consistency between predicted progress and execution order.

Figure~\ref{fig:visual_rewards} compares TwinBridge with ReWiND~\cite{zhang2025rewind}, LIV~\cite{ma2023liv}, and VIP~\cite{ma2023vip} on the same real tube-loading rollout.
The baselines use eight shared real demonstrations for 2,000 updates; the plotted rollout is excluded from training.
ReWiND uses normalized frame position $t/T$ as its demonstration progress target, with video rewinding providing supervision for progress reversals~\cite{zhang2025rewind}.
Physical task advancement need not be proportional to elapsed video time, particularly when motions pause or change speed.
TwinBridge uses simulator-derived geometric progress and task-event supervision to ground feedback in physical advancement and discrete milestones.
LIV and VIP compare each frame with demonstration endpoints in their learned feature spaces.
For the common $[0,1]$ axis, LIV uses $(1+s_{\cos})/2$ and VIP uses $\exp(-d/\tau)$, with $\tau$ fixed from the training demonstrations.
Here, $s_{\cos}$ and $d$ denote mean goal cosine similarity and feature distance, respectively.
Four synchronized images identify reaching, lifting, alignment, and insertion.

TwinBridge exhibits local score decreases and recoveries as the trajectory advances through these stages.
ReWiND continues to rise overall, and its curve follows a more uniform upward trend throughout the rollout.
LIV decreases through much of the early and middle execution, while VIP varies within a relatively narrow range before both scores rise near insertion.

Beyond the scalar curves, TwinBridge jointly predicts task events and progress, providing explicit milestone evidence alongside the continuous signal.
Table~\ref{tab:visual_reward_accuracy} evaluates event recognition across the five tasks, while the blue ticks in Fig.~\ref{fig:visual_rewards} mark progress-credit updates in the ledger.
The readout separates current progress evidence from credit already assigned during the episode.
Eligible events receive task-weighted credit once after temporal confirmation in the prescribed order; progress receives additional credit only when a valid new record exceeds the previous maximum.
Consequently, a score recovery below that maximum or a return to an already credited milestone does not earn duplicate credit.
Thus, TwinBridge combines local progress feedback with milestone rewards and a ledger that prevents duplicate credit.

\section{CONCLUSIONS}
\label{sec:conclusion}

TwinBridge uses digital twins to transfer both manipulation policies and physically grounded process supervision to real-world online RL.
The shared twin links behavior acquisition and reward construction through synchronized visual observations and physical states.
Visual distillation carries event and progress evidence into a history-aware reward readout, guiding policy refinement with human interventions. Across five laboratory tasks, TwinBridge achieved a mean final autonomous success rate of 98\%.
The results show how simulation can supply initial behavior and deployable process feedback for learning precise laboratory manipulation.

\par\vspace{\baselineskip}
\noindent
\begin{minipage}{\columnwidth}
\section*{APPENDIX}
\vspace{0.5\baselineskip}
\makeatletter
\@twocolumnfalse
\makeatother
\begin{algorithm}[H]
\footnotesize
\DontPrintSemicolon
\SetAlFnt{\footnotesize}
\SetAlCapFnt{\footnotesize\bfseries}
\SetAlCapNameFnt{\footnotesize\bfseries}
\caption{TwinBridge Policy and Reward Transfer}
\label{alg:twinbridge}
\SetAlgoNlRelativeSize{-1}
\LinesNumbered
\SetInd{0.4em}{0.7em}
\KwIn{Digital twin; simulated demonstrations; real visual annotations; task reward configuration.}
\KwOut{Refined policy $\pi_\theta$; frozen visual predictor $f_\psi$.}
\textbf{Policy and reward preparation}\;
Construct the physical ledger and reward (Eqs.~\eqref{eq:credit_update}--\eqref{eq:total_reward})\;
Pretrain $\pi_\theta$ offline on simulated demonstrations\;
Improve $\pi_\theta$ by simulation RL; collect RGB sequences and physical event/progress targets\;
Train $f_\psi$ on these targets and real visual annotations with task-specific weights (Eq.~\eqref{eq:visual_loss}); freeze $\psi$\;
Prepare the transferred policy, real critics, and replay buffers\;
\textbf{Real interaction (concurrent with learner)}\;
\For{each real episode}{
  Reset observation, recurrent state, and ledger; $t\gets0$, $C_0\gets0$, $\widehat\Phi_0\gets0$, $\tau\gets\emptyset$\;
  \While{episode is active}{
    Execute the policy command or human override; obtain executed $a_t$ and observation $o_{t+1}$\;
    Infer visual events/progress with frozen $f_\psi$; advance its recurrent state\;
    Confirm ordered events ($p\geq0.7$ for two consecutive observations); resolve the task outcome and end flag $d_t$\;
    \eIf{$d_t=1$}{
      $\widehat\Phi_{t+1}\gets0$; retain prior event/progress credit\;
    }{
      Update new-event and valid progress-record credit by Eq.~\eqref{eq:credit_update}; compute $\widehat\Phi_{t+1}$ by Eq.~\eqref{eq:ledger_potential}\;
    }
    $r_t\gets r_t^{\mathrm{base}}+\beta(\gamma\widehat\Phi_{t+1}-\widehat\Phi_t)+r_t^{\mathrm{aux}}$\;
    Record $(o_t,a_t,r_t,o_{t+1},1-d_t)$, intervention status, and critic ledger states in $\tau$; advance $t$\;
  }
  Finalize the outcome, terminal reward, and bootstrap mask; submit recorded transitions to replay\;
}
\textbf{Concurrent learner}\;
Wait until replay is ready\;
\While{training continues}{
  Sample a batch from the replay buffers\;
  Optimize the actor and critics with the HIL-ConRFT objective (Sec.~\ref{sec:online_optimization}); publish the policy\;
}
\end{algorithm}

\end{minipage}
\par

\bibliographystyle{IEEEtran}
\bibliography{root}

\begin{thebibliography}{10}
\providecommand{\url}[1]{#1}
\csname url@rmstyle\endcsname
\providecommand{\newblock}{\relax}
\providecommand{\bibinfo}[2]{#2}
\providecommand\BIBentrySTDinterwordspacing{\spaceskip=0pt\relax}
\providecommand\BIBentryALTinterwordstretchfactor{4}
\providecommand\BIBentryALTinterwordspacing{\spaceskip=\fontdimen2\font plus
\BIBentryALTinterwordstretchfactor\fontdimen3\font minus
  \fontdimen4\font\relax}
\providecommand\BIBforeignlanguage[2]{{%
\expandafter\ifx\csname l@#1\endcsname\relax
\typeout{** WARNING: IEEEtran.bst: No hyphenation pattern has been}%
\typeout{** loaded for the language `#1'. Using the pattern for}%
\typeout{** the default language instead.}%
\else
\language=\csname l@#1\endcsname
\fi
#2}}

\bibitem{luo2025precise}
J.~Luo, C.~Xu, J.~Wu, and S.~Levine, ``Precise and dexterous robotic
  manipulation via human-in-the-loop reinforcement learning,'' \emph{Sci.
  Robot.}, vol.~10, no. 105, 2025, {Art. no. eads5033, doi:
  \url{10.1126/scirobotics.ads5033}}.

\bibitem{chen2025conrft}
Y.~Chen, S.~Tian, S.~Liu, Y.~Zhou, H.~Li, and D.~Zhao, ``{ConRFT}: A reinforced
  fine-tuning method for {VLA} models via consistency policy,'' in \emph{Proc.
  Robot. Sci. Syst. (RSS)}, 2025, {doi: \url{10.15607/RSS.2025.XXI.019}}.

\bibitem{qureshi2025splatsim}
M.~N. Qureshi, S.~Garg, F.~Yandun, D.~Held, G.~Kantor, and A.~Silwal,
  ``{SplatSim}: Zero-shot {Sim2Real} transfer of {RGB} manipulation policies
  using {Gaussian} splatting,'' in \emph{Proc. IEEE Int. Conf. Robot. Autom.
  (ICRA)}, 2025, pp. 6502--6509, {doi: \url{10.1109/ICRA55743.2025.11128339}}.

\bibitem{jiang2026gsworld}
G.~Jiang, \emph{et~al.}, ``{GSWorld}: Closed-loop photorealistic simulation
  suite for robot manipulation,'' in \emph{Proc. IEEE Int. Conf. Robot. Autom.
  (ICRA)}, 2026, pp. 384--391, {doi: \url{10.1109/ICRA57385.2026.11697084}}.

\bibitem{wu2025simlauncher}
M.~Wu, \emph{et~al.}, ``{SimLauncher}: Launching sample-efficient real-world
  robotic reinforcement learning via simulation pre-training,'' in \emph{Proc.
  IEEE/RSJ Int. Conf. Intell. Robots Syst. (IROS)}, 2025, pp. 7933--7940, {doi:
  \url{10.1109/IROS60139.2025.11246668}}.

\bibitem{xu2026twinrl}
Q.~Xu, \emph{et~al.}, ``{TwinRL}: Digital twin-driven reinforcement learning
  for real-world robotic manipulation,'' \emph{arXiv preprint
  arXiv:2602.09023}, 2026.

\bibitem{chen2026sarm}
Q.~Chen, J.~Yu, M.~Schwager, P.~Abbeel, F.~Shentu, and P.~Wu, ``{SARM}:
  Stage-aware reward modeling for long horizon robot manipulation,'' in
  \emph{Proc. Int. Conf. Learn. Represent. (ICLR)}, 2026, pp. 17\,091--17\,116.

\bibitem{tan2026robodopamine}
H.~Tan, \emph{et~al.}, ``General process reward modeling for robotic
  reinforcement learning,'' in \emph{Proc. IEEE/CVF Conf. Comput. Vis. Pattern
  Recognit. (CVPR)}, 2026, pp. 22\,412--22\,422.

\bibitem{qiu2026keyframe}
Y.~Qiu, S.~Sun, H.~Ye, R.~X. Xu, and M.~Sun, ``Keyframe-guided structured
  rewards for reinforcement learning in long-horizon laboratory robotics,''
  \emph{arXiv preprint arXiv:2603.00719}, 2026.

\bibitem{burger2020chemist}
B.~Burger, \emph{et~al.}, ``A mobile robotic chemist,'' \emph{Nature}, vol.
  583, no. 7815, pp. 237--241, 2020, {doi: \url{10.1038/s41586-020-2442-2}}.

\bibitem{dai2024autonomous}
T.~Dai, \emph{et~al.}, ``Autonomous mobile robots for exploratory synthetic
  chemistry,'' \emph{Nature}, vol. 635, no. 8040, pp. 890--897, 2024, {doi:
  \url{10.1038/s41586-024-08173-7}}.

\bibitem{boiko2023autonomous}
D.~A. Boiko, R.~MacKnight, B.~Kline, and G.~Gomes, ``Autonomous chemical
  research with large language models,'' \emph{Nature}, vol. 624, no. 7992, pp.
  570--578, 2023, {doi: \url{10.1038/s41586-023-06792-0}}.

\bibitem{qiu2025biomars}
Y.~Qiu, \emph{et~al.}, ``{BioMARS}: A multi-agent robotic system for autonomous
  biological experiments,'' \emph{arXiv preprint arXiv:2507.01485}, 2025.

\bibitem{ibarz2021robot}
J.~Ibarz, J.~Tan, C.~Finn, M.~Kalakrishnan, P.~Pastor, and S.~Levine, ``How to
  train your robot with deep reinforcement learning: lessons we have learned,''
  \emph{Int. J. Robot. Res.}, vol.~40, no. 4--5, pp. 698--721, 2021, {doi:
  \url{10.1177/0278364920987859}}.

\bibitem{kalashnikov2018qtopt}
D.~Kalashnikov, \emph{et~al.}, ``Scalable deep reinforcement learning for
  vision-based robotic manipulation,'' in \emph{Proc. Conf. Robot Learn.
  (CoRL)}, 2018, pp. 651--673.

\bibitem{ball2023rlpd}
P.~J. Ball, L.~Smith, I.~Kostrikov, and S.~Levine, ``Efficient online
  reinforcement learning with offline data,'' in \emph{Proc. Int. Conf. Mach.
  Learn. (ICML)}, 2023, pp. 1577--1594.

\bibitem{hu2024ibrl}
H.~Hu, S.~Mirchandani, and D.~Sadigh, ``Imitation bootstrapped reinforcement
  learning,'' in \emph{Proc. Robot. Sci. Syst. (RSS)}, 2024, {doi:
  \url{10.15607/RSS.2024.XX.056}}.

\bibitem{kelly2019hgdagger}
M.~Kelly, C.~Sidrane, K.~Driggs-Campbell, and M.~J. Kochenderfer,
  ``{HG-DAgger}: Interactive imitation learning with human experts,'' in
  \emph{Proc. IEEE Int. Conf. Robot. Autom. (ICRA)}, 2019, pp. 8077--8083,
  {doi: \url{10.1109/ICRA.2019.8793698}}.

\bibitem{qiu2026graft}
Y.~Qiu, H.~Ye, S.~Sun, Z.~Huang, R.~X. Xu, and M.~Sun, ``{GRAFT}: Grounded and
  efficient online reinforcement adaptation for fine-grained robot
  manipulation,'' \emph{arXiv preprint arXiv:2608.27079}, 2026.

\bibitem{tobin2017randomization}
J.~Tobin, R.~Fong, A.~Ray, J.~Schneider, W.~Zaremba, and P.~Abbeel, ``Domain
  randomization for transferring deep neural networks from simulation to the
  real world,'' in \emph{Proc. IEEE/RSJ Int. Conf. Intell. Robots Syst.
  (IROS)}, 2017, pp. 23--30, {doi: \url{10.1109/IROS.2017.8202133}}.

\bibitem{peng2018dynamics}
X.~B. Peng, M.~Andrychowicz, W.~Zaremba, and P.~Abbeel, ``Sim-to-real transfer
  of robotic control with dynamics randomization,'' in \emph{Proc. IEEE Int.
  Conf. Robot. Autom. (ICRA)}, 2018, pp. 3803--3810, {doi:
  \url{10.1109/ICRA.2018.8460528}}.

\bibitem{torne2024rialto}
M.~Torne~Villasevil, \emph{et~al.}, ``Reconciling reality through simulation: A
  real-to-sim-to-real approach for robust manipulation,'' in \emph{Proc. Robot.
  Sci. Syst. (RSS)}, 2024, {doi: \url{10.15607/RSS.2024.XX.015}}.

\bibitem{han2026re3sim}
X.~Han, \emph{et~al.}, ``{RE$^3$SIM}: Generating high-fidelity simulation data
  via {3D}-photorealistic real-to-sim for robotic manipulation,'' in
  \emph{Proc. IEEE Int. Conf. Robot. Autom. (ICRA)}, 2026, pp.
  20\,781--20\,788, {doi: \url{10.1109/ICRA57385.2026.11695858}}.

\bibitem{wu2025rlgsbridge}
Y.~Wu, \emph{et~al.}, ``{RL-GSBridge}: {3D} {Gaussian} splatting based
  {Real2Sim2Real} method for robotic manipulation learning,'' in \emph{Proc.
  IEEE Int. Conf. Robot. Autom. (ICRA)}, 2025, pp. 192--198, {doi:
  \url{10.1109/ICRA55743.2025.11128103}}.

\bibitem{fan2025twinaligner}
H.~Fan, \emph{et~al.}, ``{TwinAligner}: Visual-dynamic alignment empowers
  physics-aware {Real2Sim2Real} for robotic manipulation,'' \emph{arXiv
  preprint arXiv:2512.19390}, 2025.

\bibitem{jiang2025transic}
Y.~Jiang, C.~Wang, R.~Zhang, J.~Wu, and L.~Fei-Fei, ``{TRANSIC}: Sim-to-real
  policy transfer by learning from online correction,'' in \emph{Proc. Conf.
  Robot Learn. (CoRL)}, 2025, pp. 1691--1729.

\bibitem{yin2025sgft}
P.~Yin, T.~Westenbroek, C.-A. Cheng, A.~Kolobov, and A.~Gupta, ``Rapidly
  adapting policies to the real-world via simulation-guided fine-tuning,'' in
  \emph{Proc. Int. Conf. Learn. Represent. (ICLR)}, 2025, pp. 92\,257--92\,278.

\bibitem{donahue2024progress}
G.~Donahue and E.~Elhamifar, ``Learning to predict activity progress by
  self-supervised video alignment,'' in \emph{Proc. IEEE/CVF Conf. Comput. Vis.
  Pattern Recognit. (CVPR)}, 2024, pp. 18\,667--18\,677, {doi:
  \url{10.1109/CVPR52733.2024.01766}}.

\bibitem{zakka2022xirl}
K.~Zakka, A.~Zeng, P.~Florence, J.~Tompson, J.~Bohg, and D.~Dwibedi, ``{XIRL}:
  Cross-embodiment inverse reinforcement learning,'' in \emph{Proc. Conf. Robot
  Learn. (CoRL)}, 2022, pp. 537--546.

\bibitem{ma2023vip}
\BIBentryALTinterwordspacing
Y.~J. Ma, S.~Sodhani, D.~Jayaraman, O.~Bastani, V.~Kumar, and A.~Zhang,
  ``{VIP}: Towards universal visual reward and representation via
  value-implicit pre-training,'' in \emph{Proc. Int. Conf. Learn. Represent.
  (ICLR)}, 2023. [Online]. Available:
  \url{https://openreview.net/forum?id=YJ7o2wetJ2}
\BIBentrySTDinterwordspacing

\bibitem{ma2023liv}
Y.~J. Ma, V.~Kumar, A.~Zhang, O.~Bastani, and D.~Jayaraman, ``{LIV}:
  Language-image representations and rewards for robotic control,'' in
  \emph{Proc. Int. Conf. Mach. Learn. (ICML)}, 2023, pp. 23\,301--23\,320.

\bibitem{wang2024rlvlmf}
Y.~Wang, \emph{et~al.}, ``{RL-VLM-F}: Reinforcement learning from vision
  language foundation model feedback,'' in \emph{Proc. Int. Conf. Mach. Learn.
  (ICML)}, 2024, pp. 51\,484--51\,501.

\bibitem{lee2026roboreward}
T.~Lee, A.~Wagenmaker, K.~Pertsch, P.~Liang, S.~Levine, and C.~Finn,
  ``{RoboReward}: General-purpose vision-language reward models for robotics,''
  \emph{arXiv preprint arXiv:2601.00675}, 2026.

\bibitem{zhang2025rewind}
J.~Zhang, \emph{et~al.}, ``{ReWiND}: Language-guided rewards teach robot
  policies without new demonstrations,'' in \emph{Proc. Conf. Robot Learn.
  (CoRL)}, 2025, pp. 460--488.

\bibitem{liu2026timerewarder}
Y.~Liu, C.~Wen, Y.~Hu, D.~Jayaraman, and Y.~Gao, ``{TimeRewarder}: Learning
  dense reward from passive videos via frame-wise temporal distance,'' in
  \emph{Proc. Int. Conf. Mach. Learn. (ICML)}, 2026, pp. 79\,029--79\,049.

\bibitem{icarte2018rewardmachines}
R.~Toro~Icarte, T.~Q. Klassen, R.~Valenzano, and S.~A. McIlraith, ``Using
  reward machines for high-level task specification and decomposition in
  reinforcement learning,'' in \emph{Proc. Int. Conf. Mach. Learn. (ICML)},
  2018, pp. 2107--2116.

\bibitem{feng2026procvlm}
Y.~Feng, \emph{et~al.}, ``{ProcVLM}: Learning procedure-grounded progress
  rewards for robotic manipulation,'' \emph{arXiv preprint arXiv:2605.08774},
  2026.

\bibitem{ng1999shaping}
A.~Y. Ng, D.~Harada, and S.~Russell, ``Policy invariance under reward
  transformations: Theory and application to reward shaping,'' in \emph{Proc.
  Int. Conf. Mach. Learn. (ICML)}, 1999, pp. 278--287.

\bibitem{kerbl2023gaussian}
B.~Kerbl, G.~Kopanas, T.~Leimk{\"u}hler, and G.~Drettakis, ``{3D} {Gaussian}
  splatting for real-time radiance field rendering,'' \emph{ACM Trans. Graph.},
  vol.~42, no.~4, pp. 139:1--139:14, 2023, {doi: \url{10.1145/3592433}}.

\bibitem{devlin2012dynamic}
S.~Devlin and D.~Kudenko, ``Dynamic potential-based reward shaping,'' in
  \emph{Proc. Int. Conf. Auton. Agents Multiagent Syst. (AAMAS)}, 2012, pp.
  433--440, {doi: \url{10.65109/JJTT8551}}.

\bibitem{pardo2018timelimits}
F.~Pardo, A.~Tavakoli, V.~Levdik, and P.~Kormushev, ``Time limits in
  reinforcement learning,'' in \emph{Proc. Int. Conf. Mach. Learn. (ICML)},
  2018, pp. 4045--4054.

\bibitem{team2024octo}
{Octo Model Team}, \emph{et~al.}, ``{Octo}: An open-source generalist robot
  policy,'' in \emph{Proc. Robot. Sci. Syst. (RSS)}, 2024, {doi:
  \url{10.15607/RSS.2024.XX.090}}.

\bibitem{cho2014gru}
K.~Cho, \emph{et~al.}, ``Learning phrase representations using {RNN}
  encoder--decoder for statistical machine translation,'' in \emph{Proc. Conf.
  Empirical Methods Nat. Lang. Process. (EMNLP)}, 2014, pp. 1724--1734, {doi:
  \url{10.3115/v1/D14-1179}}.

\bibitem{song2023consistency}
Y.~Song, P.~Dhariwal, M.~Chen, and I.~Sutskever, ``Consistency models,'' in
  \emph{Proc. Int. Conf. Mach. Learn. (ICML)}, 2023, pp. 32\,211--32\,252.

\end{thebibliography}
\end{CJK*}
\end{document}